# Embedding-Enhanced Probabilistic Modeling of Ferroelectric Field Effect Transistors (FeFETs)

Tasnia Nobi Afee[1], Jack Hutchins[2], Md Mazharul Islam[1], Thomas Kämpfe[3], and Ahmedullah Aziz[1*]

[1] Dept. of Electrical Eng. and Computer Sci., University of Tennessee, Knoxville, TN, 37996, USA
[2] Oak Ridge National Laboratory, Oak Ridge, TN 37831, USA
[3] Fraunhofer IPMS, Center Nanoelectronic Technologies (CNT), Dresden, Germany
[*] Corresponding Author. Email: aziz@utk.edu

**Abstract:** FeFETs hold strong potential for advancing memory and logic technologies, but their inherent randomness, arising from both operational cycling and fabrication variability, poses significant challenges for accurate and reliable modeling. Capturing this variability is critical, as it enables designers to predict behavior, optimize performance, and ensure reliability and robustness against variations in manufacturing and operating conditions [1]. Existing deterministic and machine learning-based compact models often fail to capture the full extent of this variability or lack the mathematical smoothness required for stable circuit-level integration. In this work, we present an enhanced probabilistic modeling framework for FeFETs that addresses these limitations. Building upon a Mixture Density Network (MDN) foundation, our approach integrates $C\infty$ continuous activation functions for smooth, stable learning and a device-specific embedding layer to capture intrinsic physical variability across devices. Sampling from the learned embedding distribution enables the generation of synthetic device instances for variability-aware simulation. With an $R^2$ of 0.92, the model demonstrates high accuracy in capturing the variability of FeFET current behavior. Altogether, this framework provides a scalable, data-driven solution for modeling the full stochastic behavior of FeFETs and offers a strong foundation for future compact model development and circuit simulation integration.

## 1. Introduction

Ferroelectric Field-Effect Transistors (FeFETs) have gained considerable attention as promising candidates for future memory and computing technologies due to their non-volatility, fast switching, low power consumption, and compatibility with standard CMOS processes. Their unique properties make them suitable for a range of emerging applications, including embedded non-volatile memory, neuromorphic computing, and in-memory logic. In addition, their ability to operate efficiently at high speeds and low energy levels makes them attractive for next-generation architectures that demand both high performance and energy efficiency [2].

However, despite their promising features, effectively harnessing FeFETs in real-world applications remains challenging due to their inherent random characteristics. Variability from cycle-to-cycle (C2C) switching and device-to-device (D2D) fabrication differences introduces significant uncertainty, complicating efforts to ensure consistent performance, design predictability, and long-term reliability. As devices scale down to nanometer dimensions, quantum mechanical effects become more significant, making it difficult for classical physics equations to accurately capture these phenomena [3]. As a result, accurately modeling this stochastic behavior is not only essential but also non-trivial.

To address such variability, machine learning-based compact models are widely used in circuit design. These models offer improvements over traditional models like BSIM and table-based methods, offering better convergence, more accurate predictions, and significantly faster simulations (82 to 308 times faster) in tools like Cadence SPECTRE compared to table models derived from the same device [4]. They provide simplified, computationally efficient representations of complex physical device behavior, enabling accurate simulations within electronic design automation (EDA) tools like SPICE. Compact models remain essential for predicting device behavior under a variety of circuit-level conditions. Although traditional physics-based compact models have incorporated stochastic properties, they often rely on fixed assumptions and device-specific tuning. In contrast, data-driven approaches such as machine learning offer a flexible framework for capturing inherent stochasticity directly from measurements, which is essential for modeling modern nanoelectronic devices under real-world variability [5, 6].

In recent years, machine learning-based methods have been increasingly explored for compact modeling due to their ability to learn directly from data, offering advantages such as reduced development time and minimal reliance on detailed physical equations. One such study has proposed an artificial neural network (ANN)-based compact modeling methodology for advanced transistors, aiming to improve Design-Technology Co-Optimization (DTCO) and improve pathfinding efficiency [7]. Neural network have also been proposed as a faster alternative to conventional compact models for device like FinFET and NC-FinFET [4]. However, many of these frameworks fall short in addressing the full spectrum of stochastic behavior inherent in modern nanoelectronic devices. For example, Zhang et al. demonstrated the use of ANNs for transistor modeling but noted that their method required extensive domain expertise and manual tuning to optimize network structure and training data [8]. Similarly, Hutchins et al. modeled an HfOx memristor using a neural network, but such approaches have not yet been applied to FeFETs [9].

In contrast, traditional physics-based compact models attempt to incorporate randomness. However, in some cases, the underlying physics may be too complex or not yet fully understood, making it difficult to develop efficient physics-based compact models. This challenge is further amplified in devices with multiple operational states, such as memristors [10], ferroelectrics [11], antiferroelectric [12], and threshold switches [13]. The presence of hysteresis and variability effects in these devices makes accurate modeling even more difficult.These often rely on fixed statistical assumptions and require extensive device-specific tuning, which limits flexibility and scalability. For example, one such model developed for Schottky barrier and reconfigurable field-effect transistors employs simplifying assumptions and fitting parameters that compromise accuracy under specific conditions, such as high channel resistance line [14]. Similarly, another physics-based explicit compact model for double-gate Reconfigurable Field-Effect Transistors (RFETs), though compatible with SPICE simulations, performs poorly under extreme bias conditions or unconventional device geometries [15].

While machine learning-based compact modeling approaches have demonstrated significant improvements in FeFET transistor modeling, they still face limitation such as overfitting, reliance on extensive training data, and limited generalizability beyond trained data ranges. Many still rely on manual tuning to optimize ANN size or struggle with scalability due to computational inefficiencies, making them impractical for large-scale FeFET circuit simulations. Additionally, several models often lack the mathematical smoothness required for seamless integration into circuit simulators, as they do not enforce $C\infty$ continuity. A $C\infty$ function is infinitely differentiable, providing the smoothness needed to ensure numerical stability and convergence during simulation. Without this property, models can exhibit abrupt transitions or irregularities, leading to solver errors and inaccurate circuit behavior predictions.

Physics-based compact models for FeFETs also rely on simplified assumptions, which may not hold under high variability or non-ideal conditions, limiting their accuracy and adaptability. To overcome these limitations, we aim to develop a probabilistic modeling framework designed to be compatible with future compact model integration and capable of capturing the full stochastic behavior of FeFETs.

In contrast to the earlier work that applied a probabilistic modeling approach to the heater cryotron, a superconducting nanowire-based device exhibiting gate-current-controlled stochastic switching behavior at cryogenic temperatures, our study focuses on Ferroelectric Field-Effect Transistors (FeFETs), which present distinct stochastic characteristics and operate under entirely different physical principles and environmental conditions [13]. While that model, based on Mixture Density Networks (MDNs), demonstrate only cycle-to-cycle (C2C) variability, it lacked mechanisms for modeling device-to-device (D2D) differences. We initially applied the same MDN framework to FeFET data and found it capable of capturing essential stochastic trends, despite differences in underlying device physics.

Building on this foundation, our model introduces two key enhancements. First, the explicit use of $C\infty$-continuous activation functions throughout the model, which ensures both smooth learning and simulation compatibility, an aspect that was overlooked in prior work. Second, we address D2D variability by incorporating a device-specific embedding layer that learns a compact vector representation of each device's unique characteristics, such as subtle differences in thickness, width, or switching behavior. An embedding layer is a trainable component used in machine learning to convert discrete inputs, such as device IDs, into meaningful continuous vector representations [16]. This allows the model to encode device-specific variability directly from data. These embeddings improve generalization across devices during training and provide a foundation for sampling realistic synthetic devices. Once trained, the embedding space can be analyzed to simulate a broad range of device behaviors with improved robustness and accuracy. Beyond simulation, such models offer several practical benefits: they can support the development of variation-aware compact models, enable synthetic data generation for training or validation, and provide valuable insight into stochastic device behavior that is often difficult to capture through analytical expressions.

Building on a successfully demonstrated probabilistic modeling framework, these enhancements extend its applicability to FeFETs and position it as a strong foundation for future compact model development.

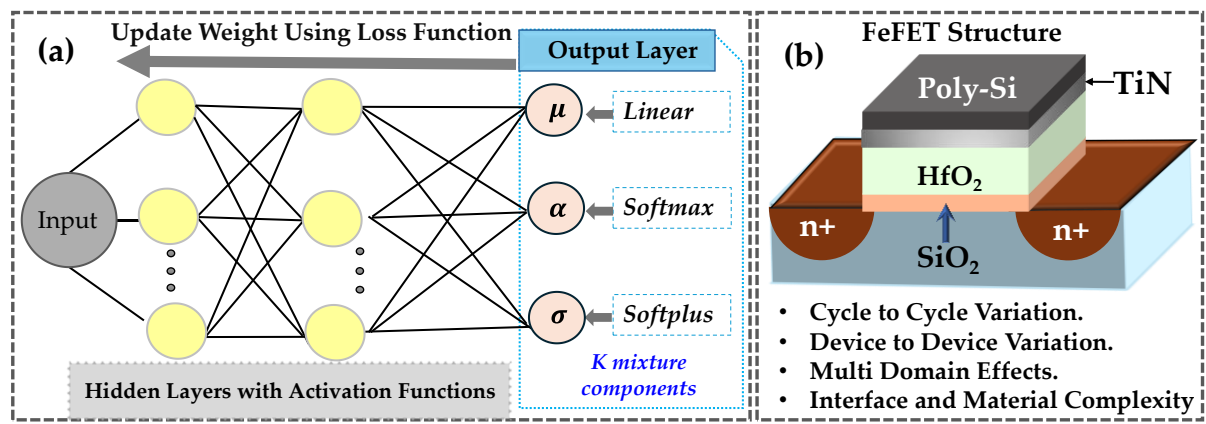

**Figure 1:(a)** Mixture Density Network (MDN) architecture for modeling stochastic behavior. **(b)** FeFET structure highlighting key layers and modeling challenges such as C2C and D2D variation.

## 2. Background

### 2.1 FeFET Structure and Key Device Features

A transistor is a fundamental building block of modern electronics, functioning as a switch or amplifier in circuits. Among various types, the Metal-Oxide-Semiconductor Field-Effect Transistor (MOSFET) is widely used due to its scalability, low power consumption, and high switching speed. A MOSFET controls current flow using an electric field applied to the gate terminal, which modulates the conductivity of a semiconductor channel. A Ferroelectric Field-Effect Transistor (FeFET) can be described as a conventional MOSFET in which the gate insulator is replaced with a ferroelectric oxide typically hafnium oxide ($HfO_2$) instead of a traditional dielectric material [17].

In the FeFET structure, as illustrated in Fig. 1(b), a ferroelectric $HfO_2$ layer is sandwiched between a metal gate (TiN) and a polycrystalline silicon (Poly-Si) gate electrode, placed over a silicon dioxide ($SiO_2$) insulator and n+ source/drain regions. This unique structure allows FeFETs to retain their polarization state even when power is removed, offering non-volatility along with the inherent advantages of MOSFETs, such as low power, fast switching, and CMOS compatibility. However, the integration of ferroelectric materials introduces new modeling challenges. Notably, Cycle-to-Cycle (C2C) variation, where the behavior of the device changes across repeated program/erase cycles, and Device-to-Device (D2D) variation, caused by manufacturing inconsistencies, significantly impact reliability and predictability. Additionally, multi-domain effects, interface states, and material complexity at the ferroelectric-semiconductor boundary further complicate accurate modeling. These stochastic behaviors and nonlinearities are difficult to capture with conventional compact models based on deterministic physics equations, highlighting the need for machine learning-based approaches that can learn these variations directly from data.

## 2.2 Mixture Density Network Architecture

The architecture of the Mixture Density Network (MDN) used in this paper is illustrated in Fig. 1(a). Unlike traditional neural networks that predict a single deterministic output, an MDN predicts the parameters of a probabilistic mixture model typically a Gaussian mixture. For each of the K number of mixture components, the MDN outputs a mean ($\mu_k$), a standard deviation ($\sigma_k$), and a mixing coefficient ($\alpha_k$) The output layer thus contains K×3 units. To ensure valid probability distributions, specific activation functions are applied: a softmax activation is used for $\alpha_k$ to guarantee non-negativity and that the mixing coefficients sum to 1 in Eq. (1).

$$s(\alpha_i) = \frac{e^{\alpha_i}}{\sum_{j=1}^{N} e^{\alpha_j}} \quad (1)$$

An exponential or softplus activation ensures $\sigma_k > 0$; and a linear activation is used for $\mu_k$ This design enables the MDN to output a complete probability density function (PDF), as shown in Eq. 2, capturing both uncertainty and multimodality in the data making it particularly effective for modeling stochastic or non-deterministic systems such as FeFET devices.

$$p(x) = \sum_{k=1}^{N} \alpha_k \cdot \frac{1}{\sqrt{2\pi\sigma_k^2}} \cdot \exp\left(-\frac{(x-\mu_k)^2}{2\sigma_k^2}\right) \quad (2)$$

The MDN is trained using the loss function, where it is computed by comparing the predicted mixture distribution to the true target values. The network typically includes a few fully connected hidden layers with nonlinear activation functions such as ReLU, tanh, or Mish. The choice of the number and size of hidden layers is guided by background knowledge and experimentation [18]. More complex tasks may benefit from deeper architectures [19]. During training, standard backpropagation is used to optimize the network weights, adjusting the predicted distribution to better fit the observed data. This probabilistic framework makes MDNs highly suitable for modeling outputs with inherent randomness or multiple plausible outcomes.

## 2.3 Prior Mixture Density Networks (MDNs) Based Modeling Framework

We build upon a previously established modeling framework that demonstrated the use of Mixture Density Networks (MDNs) to capture stochastic behavior in memristive devices such as heater cryotrons [20]. However, despite its success in modeling randomness in memristors, this MDN-based approach has not yet been explored for FeFETs, which exhibit distinct variability characteristics.

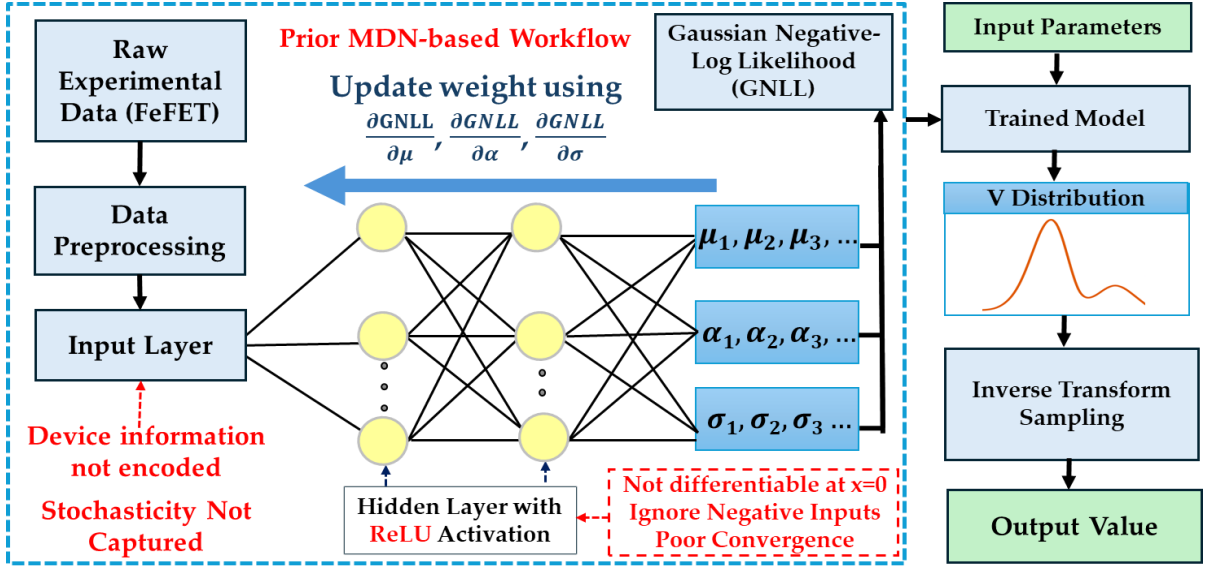

**Figure 2:** Prior MDN-based modelling workflow

As shown in Fig. 2, this model takes experimental data as input and applies a standard MDN architecture using ReLU activation and Gaussian Negative Log-Likelihood (GNLL) loss for training. However, this framework presents two key limitations when applied to FeFETs. First, it does not capture the stochasticity in the device characteristics, limiting its ability to model device-to-device (D2D) variation. Second, the use of ReLU activation, which is non-differentiable at zero and can lead to gradient issues for negative inputs, may hinder smooth convergence and integration into circuit simulation tools that require differentiable models.

To address these challenges, our baseline model introduces smooth $C\infty$-continuous activation functions (Mish) to ensure compatibility with compact modeling environments, as well as other enhancements aimed at extending applicability to FeFET devices.

## 3. Methods

### 3.1 C∞ Continuity Enforcement for Stable Simulation

One of the key novelties of our model is the enforcement of C-infinity (C∞) continuity, a critical mathematical property that ensures the function and all its derivatives are smooth and continuous at all orders. This level of smoothness is not just a technical preference,it fundamentally enhances both training stability and compatibility with modern circuit simulators by eliminating abrupt changes that could hinder optimization or disrupt convergence.

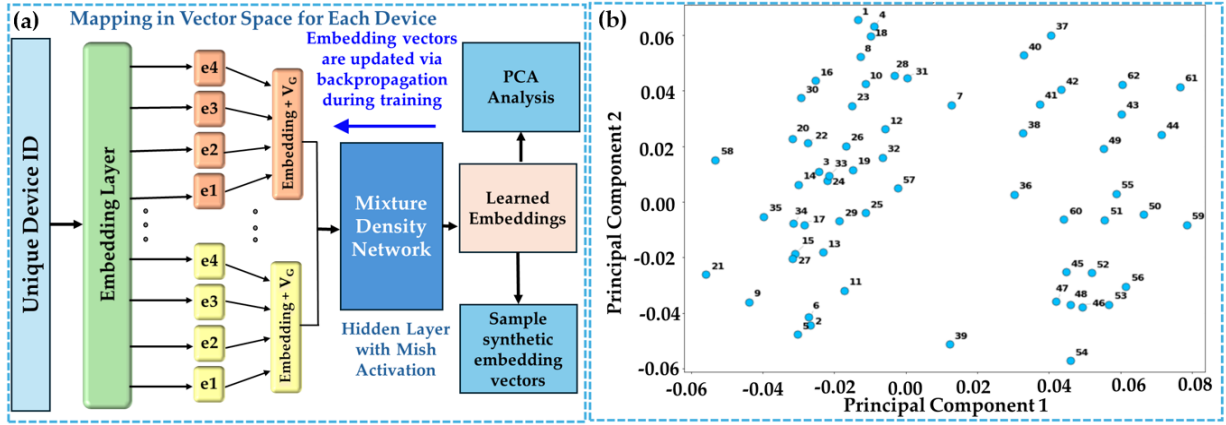

**Figure 3: (a)** Overview of device-specific embedding integration for modeling device-to-device variability. **(b)** PCA plot of learned embeddings where each number representing a specific FeFET device (Unique device ID 0–62). The clustering illustrates how the model groups devices based on similarities in their learned I–V characteristics.

Prior works often relied on activation functions like ReLU, valued for their simplicity and efficiency in standard deep learning. However, ReLU is only C⁰ continuous, meaning it is not differentiable at zero and exhibits sharp transitions in its first derivative. While tolerable in conventional applications, these discontinuities pose serious challenges in compact modeling, including poor convergence, unstable behavior, and reduced simulation accuracy, particularly when integrated into SPICE-based environments.

To address this, we replace ReLU with the Mish activation function, which has been analytically proven to be C∞ continuous [2] and allows for smooth gradient flow throughout the model. This ensures our model behaves reliably during both training and simulation. This design decision is grounded in the best practices for compact modeling, as emphasized by McAndrew et al. [15], who clearly state that models intended for circuit simulation must use smooth and differentiable equations, ideally with C∞ continuity, to guarantee reliable convergence with the nonlinear solvers used in SPICE and related tools. The Mish activation is defined as shown in Eq. (3.3):

$$\text{Mish}(x) = x \cdot \tanh(\text{softplus}(x)) \qquad (3)$$

$$\text{where, softplus}(x) = \ln(1 + e^x)$$

By incorporating the Mish function and enforcing C∞ continuity throughout our model, we meet and exceed the rigorous requirements for numerical smoothness in circuit simulation. This not only improves convergence and numerical stability but also ensures our model is seamlessly compatible with modern simulation environments. Thus, the adoption of this smoothness property is not merely a technical refinement, it represents a key innovation and a foundational element of our modeling strategy.

## 3.2 Integration of Embedding Layers for D2D Variability

To explicitly address device-to-device (D2D) variability in FeFET behavior, we integrate a device-specific embedding layer into our model architecture. This embedding layer serves as a compact, learnable representation of intrinsic device characteristics, capturing stochastic behaviors arising from process-induced factors such as layer thickness, grain orientation, or defect density without requiring explicit knowledge of these physical parameters. Rather than treating device IDs as simple categorical inputs, our model maps each device to a continuous vector in a learned embedding space, allowing it to infer and encode device-specific electrical behaviors in a differentiable and data-driven manner.

Specifically, each device ID is associated with a 4-dimensional embedding vector ($e_1$, $e_2$, $e_3$, $e_4$), which is randomly initialized and trained jointly with the rest of the network. During training, this vector is concatenated with the input gate voltage ($V_G$), forming a composite input that incorporates both electrical conditions and the intrinsic characteristics of the device. As the model learns, these embeddings are refined via backpropagation, enabling the model to accurately capture variations in *I–V* behavior across devices. The complete workflow is depicted in Fig. 3a, where each device is mapped to its own learned embedding vector that influences the stochastic output predictions.

To guide this learning process, we employ a loss function, which jointly updates both the network weights and the device-specific embedding vectors by comparing the predicted probabilistic distribution to the true observed current values. Rather than using traditional negative log-likelihood loss, we adopted the Continuous Ranked Probability Score (CRPS) as the loss function to better evaluate the accuracy of the predicted cumulative distribution function (CDF) against the true value. The CRPS is particularly useful for probabilistic regression tasks where full distributional accuracy matters. CRPS is a scoring rule used to evaluate how well a predicted distribution matches an observed outcome. Unlike typical loss functions that compare a predicted value to a true value, CRPS compares the entire predicted distribution (CDF) to the true observation as shown in Eq. (4).

$$\text{CRPS}(F, y) = E[|X - y|] - \frac{1}{2} E[|X - X'|] \qquad (4)$$

Here, X and X′ are samples from the predicted distribution F, and y is the true observed value. The first term measures the average distance between the predicted values and the actual value, while the second term captures the internal spread of the predicted distribution. Subtracting this spread helps balance accuracy with how confident or concentrated the prediction is. A lower CRPS indicates that the predicted distribution is both accurate and appropriately focused around the true value

During training, the model utilizes backpropagation method [21] and adjusts its internal weights based on the CRPS score to improve predictive accuracy and update the embedding vectors. We employ Python in combination with TensorFlow [22] and Keras [23]. Keras facilitates the efficient development, training, and evaluation of neural networks, making it well-suited for our application. In our case, the model predicts a mixture of several Gaussian components, each defined by its own mean, standard deviation, and mixing coefficient. This mixture enables the model to represent complex, multimodal distributions. For such settings, CRPS has a specific form adapted for Gaussian mixture models, as shown below in Eq. (5).

$$\text{CRPS}(y) = \sum_{k=1}^{K} \alpha_k \, E[|X_k - y|] - \frac{1}{2} \sum_{k=1}^{K} \sum_{l=1}^{K} \alpha_k \alpha_l \, E[|X_k - X_l|] \qquad (5)$$

To better understand and interpret the structure of the learned embedding space, we apply Principal Component Analysis (PCA), a statistical technique used to reduce high-dimensional data into a lower-dimensional form while preserving the most significant patterns. In our case, PCA allows us to project the high-dimensional device embeddings into a two-dimensional space, making it possible to visualize relationships between devices in a way that is easy to interpret. This dimensionality reduction preserves the most critical variance, enabling a clear visualization of how the model organizes device behavior in the latent space. In the resulting PCA plot (Fig. 3b), each point represents a distinct device, labeled by its device ID. Devices that appear close together are inferred to have similar current–voltage characteristics, while those located farther apart likely exhibit different electrical responses or variability trends. Remarkably, these clusters emerge without the model being given any explicit physical information, such as geometry or thickness, suggesting that the embedding layer has successfully learned meaningful behavioral representations. To further analyze the structure, we fit a multivariate Gaussian distribution over the embedding space, offering a compact probabilistic interpretation. This framework enables generalization across devices, supports synthetic device generation, and facilitates variability-aware circuit simulation, all from a single trained model.

Fig. 4 illustrates the overall architecture of the proposed probabilistic modeling framework for FeFETs. It consists of two key components: a device-specific embedding layer and a Mixture Density Network (MDN) with C∞-continuous activation functions. During training, each device ID is mapped to a learnable embedding vector, which is concatenated with the input gate voltage and passed through the MDN. The MDN outputs a probabilistic distribution (mixture of Gaussians) representing the drain current. After training, the learned embeddings are used for synthetic device generation by sampling from a fitted multivariate Gaussian distribution, enabling realistic and variability-aware I–V predictions.

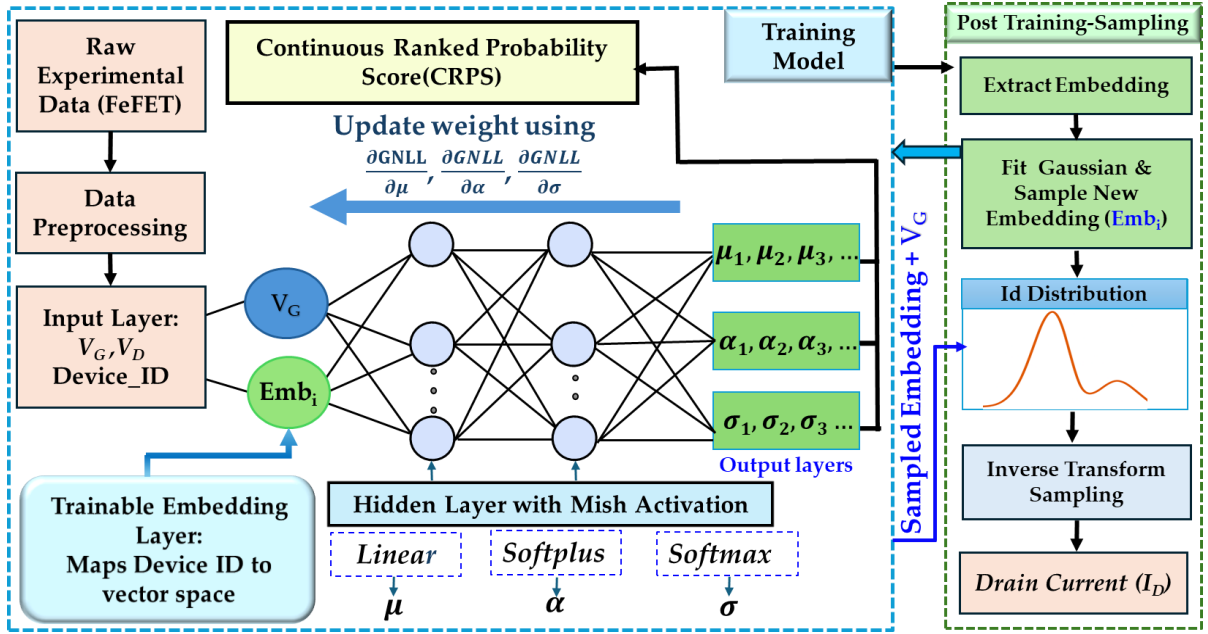

**Figure 4:** Overview of the proposed probabilistic framework for FeFETs, combining a trainable embedding layer and C∞-continuous activations to capture both device-to-device and cycle-to-cycle variations. Learned embeddings enable synthetic device generation for variability-aware circuit simulations

## 4.2 Synthetic Device Generation via Embedding Sampling

Using the proposed embedding-enhanced MDN architecture, we first learned a low-dimensional representation of device-specific variability. During training, each device is assigned a unique embedding vector that is optimized along with the network weights to reflect its distinct current-voltage (*I–V*) characteristics. As a result, the model internalizes process-dependent features in a latent embedding space, effectively learning a low-dimensional manifold of device variability. After training, we extract the learned embedding vectors for all devices and fit a multivariate Gaussian distribution to these vectors expressed as Eq. (5).

$$e \sim N(\mu_{emb}, \Sigma_{emb})  \quad (5)$$

where $\mu_{emb}$ and $\mu_{emb}$ and $\Sigma_{emb}$ are the mean and covariance of the trained embedding vectors. This probabilistic distribution serves as a generator for new, unseen device characteristics.

To synthesize new device behavior, we sample new embedding vectors $e_{new}$ from the learned embedding distribution. To ensure physically consistent and stable outputs during sampling, we use inverse transform sampling on the learned truncated Gaussian mixture model [16]. These vectors are then passed through the trained model to generate stochastic I–V characteristics. These vectors represent realistic, unseen combinations of device-specific features. Importantly, since the trained MDN model has already learned to associate embedding vectors with corresponding *I–V* output distributions, passing these new embeddings through the model produces physically plausible and statistically consistent current-voltage characteristics, even for devices do not present during training. To ensure a diverse and representative set of synthetic devices, we adopt a mixed embedding selection strategy that combines both statistical sampling and random sampling from the learned embedding distribution. Specifically, we include:

- The mean vector of the embedding distribution, representing the average device behavior.
- The mode, if distinguishable, indicating the most likely embedding configuration.
- Embeddings located ±2 standard deviations from the mean, which capture extreme but plausible variations in device behavior (e.g., edge-case scenarios).
- A set of randomly sampled embedding vectors from the multivariate Gaussian prior to explore the broader stochastic variability encoded in the embedding space.

This combination allows us to probe the embedding space systematically and stochastically, ensuring that both typical and rare behaviors are included. The sampled embeddings are then passed through the trained the model to generate corresponding synthetic I–V characteristics. These sampled embeddings can be used as inputs to circuit simulations under standardized biasing conditions. These set of predicted I–V characteristics representing diverse device behaviors.

A simple way to generate predictions from a learned distribution is to randomly sample at each time step, but this can cause abrupt, unrealistic fluctuations especially near switching points leading to unstable simulations. To avoid this, we use inverse transform sampling, which provides smooth and consistent outputs while preserving stochastic behavior. This method works by first building the cumulative distribution function (CDF) from the predicted Gaussian mixture model. A fixed quantile value $q \in [0,1]$ is sampled per device sweep and used to invert the CDF, producing temporally coherent current predictions that reflect C2C variability more accurately. To ensure physical validity, we apply truncated Gaussian sampling so that all current values remain non-negative. This is implemented using a truncated inverse CDF sampler such as truncnorm.ppf in Python.

To formally define this process, we first derive the truncated Gaussian mixture distribution and its cumulative distribution function (CDF), which are essential for implementing inverse transform sampling. The corresponding current is then computed by solving F(x)=q, is the truncated mixture CDF. The truncated normal probability density function (PDF) is derived by renormalizing the standard Gaussian PDF to account for the truncated domain, ensuring that the total probability integrates to one over the interval $[0,\infty)$. The corresponding truncated normal PDF is expressed as Eq. (6).

$$p_k^{trunc}(x) = \frac{1}{\sigma_k\sqrt{2\pi}} \cdot \frac{exp\left(-\frac{(x-\mu_k)^2}{2\sigma_k^2}\right)}{1-\Phi\left(\frac{0-\mu_k}{\sigma_k}\right)}, \quad x \geq 0 \qquad (6)$$

The predicted distribution is modelled as a weighted sum of truncated Gaussians as shown in Eq. (7).

$$p(x) = \sum_{k=1}^{K} \alpha_k \cdot \frac{1}{\sigma_k\sqrt{2\pi}} \cdot \frac{\exp\left(-\frac{(x-\mu_k)^2}{2\sigma_k^2}\right)}{1-\Phi\left(\frac{0-\mu_k}{\sigma_k}\right)}, \quad x \geq 0 \tag{7}$$

The denominator normalizes the Gaussian over the truncated domain x≥0

Once the truncated mixture PDF is defined, the CDF at point x is computed by integrating over the valid domain, as shown in Eq. (8).

$$F(x) = \int_0^x p(t)\, dt = \int_0^x \sum_{k=1}^{K} \alpha_k \cdot p_k^{\text{trunc}}(t)\, dt \tag{8}$$

To compute the full mixture CDF, we first need the CDF of each truncated component as in Eq. (9).

$$F_k^{trunc}(x) = \frac{\Phi\left(\frac{x-\mu_k}{\sigma_k}\right)-\Phi\left(\frac{0-\mu_k}{\sigma_k}\right)}{1-\Phi\left(\frac{0-\mu_k}{\sigma_k}\right)}, \quad x \geq 0 \tag{9}$$

where Φ.() is the standard normal cumulative distribution function (CDF). The normal CDF Φ(x) is computed using the error function, defined as shown in Eq. (10).

$$\Phi(x) = \frac{1}{2}\left(1 + \text{erf}\left(\frac{x}{\sqrt{2}}\right)\right) \tag{10}$$

Defining this in terms of the error function, simplifies computation. Because it is well-defined, numerically stable, and widely supported in scientific libraries. The total mixture CDF is then a weighted sum of the truncated CDFs as shown in Eq. (11).

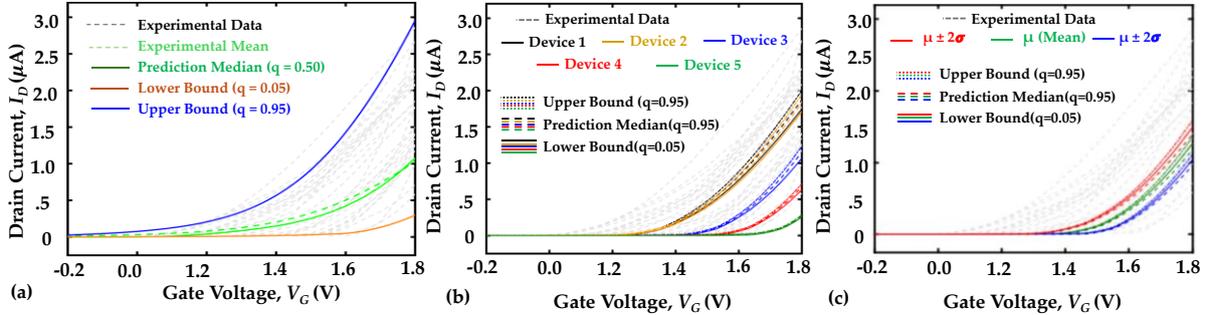

**Figure 5:** Experiment data vs Simulated I-$V$ curves for FeFET devices under three modeling configurations. **(a)** Quantile-based predictions using the baseline MDN model without device-specific embeddings. **(b)** Synthetic devices generated using randomly sampled embeddings from the learned distribution. **(c)** Using embeddings sampled from the mean and ±2 standard deviations.

$$F(x) = \sum_{k=1}^{K} \alpha_k \cdot F_k^{trunc}(x) \tag{11}$$

Since this equation has no closed-form inverse, we apply Brent's method [17], a robust numerical root-finding technique, to find the solution. To further enhance stability and prevent outlier behavior, we clip q to a range (e.g., 0.05 to 0.95), limiting predictions beyond two standard deviations. Also, a new q value is generated only when the time derivative of the input signal crosses zero a condition efficiently implemented in Verilog-A. This approach preserves the model's stochastic nature while producing stable, realistic transient simulations, making it highly suitable for compact modeling of FeFETs.

## 4. Results

Fig. 5(a) shows the model output without incorporating device-specific embedding layers. Here, the quantile-based predictions median (q=0.50) and lower (q=0.05), upper (q= 0.95) bounds are plotted across all devices. The predicted curves closely follow the experimental trends reflecting the range of current values considering a single device might exhibit over multiple cycles at each voltage point.

Fig. 5(b) presents the results after integrating a device-specific embedding layer into the MDN architecture. In this case, embeddings were randomly sampled from the learned embedding distribution, allowing the model to generate distinct I–V characteristics for synthetic devices. Fig. 5(c) shows the predictions generated from embeddings sampled systematically from the mean and ±2×standard deviations of the learned embedding space.

One challenge with Mixture Density Networks (MDNs) is the lack of intuitive numerical metrics for performance evaluation. As a result, we rely on graphical comparisons between the model's predicted output distributions and experimental measurements. Fig. 6, presents histograms of the experimental drain current ($I_D$) overlaid with the predicted probability density functions (PDFs) generated by the model at six different gate voltages: 0.9 V, 1.2 V, 1.3 V, 1.5 V, 1.6 V, and 1.7 V. The predicted PDFs are averaged over 500 synthetic devices sampled from the learned embedding distribution. On the entire dataset we also evaluate performance using traditional regression metrics and the model achieves an $R^2$ score of 0.92. Here $R^2$ value represents the percentage of variance explained by the model.

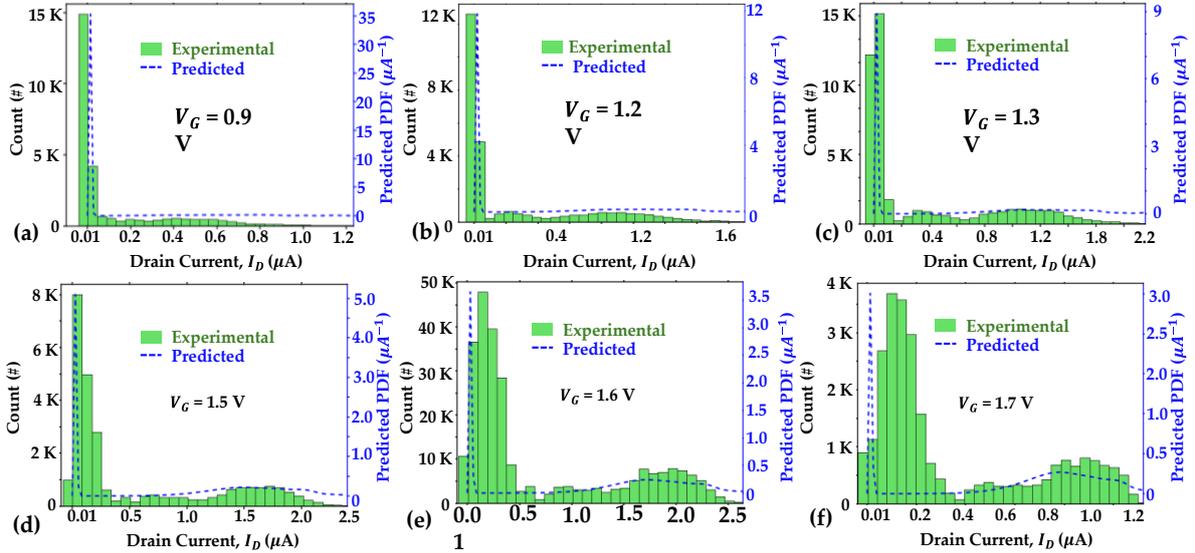

**Figure 6:** Model validation for drain current ($I_D$) against experiment devices at gate voltages of **(a)** 0.90 V, **(b)** 1.20 V, **(c)** 1.30 V, **(d)** 1.50 V, **(e)** 1.60 V, and **(f)** 1.70 V. Experimental data is shown as histograms; predicted PDFs are averaged over 500 synthetic embeddings.

## 5. Discussion

The results presented in Fig. 5 and Fig. 6 demonstrate the effectiveness of the proposed MDN-based probabilistic modeling framework. The quantile-based predictions in Fig. 5(a) offer meaningful insight into the cycle-to-cycle variability of FeFET devices, capturing the range of likely current responses under repeated operation. However, a key limitation of this approach lies in its assumption that all devices follow an identical current distribution. This simplification overlooks device-to-device (D2D) variability and limits the model's ability to distinguish between structurally or behaviorally distinct devices. Incorporating device-specific embeddings, as illustrated in Figs. 5 (b,c), further demonstrates the effectiveness of the proposed MDN-based probabilistic modeling framework in capturing both cycle-to-cycle (C2C) and device-to-device (D2D) variability in FeFET behavior.

As shown in Fig. 5(b), randomly sampling from the learned embedding space allows the model to generate synthetic devices that exhibit distinct I–V characteristics, closely matching the experimental variability across real devices. This highlights the model's ability to simulate a realistic and diverse device population, enabling exploration of unseen behaviors beyond the original training data. In Fig. 5(c), we evaluate the model using selected embeddings (mean and ±2 standard deviations) which represent nominal and edge-case device behaviors. These strategically chosen points reflect meaningful positions in the embedding distribution, simulating both typical and extreme responses based on what the model has learned, and help validate that it captures realistic device-to-device variation consistent with experimental trends. Together, these two plots show that the model generate not just a single prediction, but a distribution of device behaviors providing valuable insight into how real devices may vary in practical applications.

Since Mixture Density Networks do not inherently provide straightforward scalar performance metrics, we evaluate the model through graphical comparison with experimental data. As shown in Fig. 6, the predicted probability density functions (PDFs), generated from 500 synthetic embeddings, closely match the distributional shape of experimental data across a range of gate voltages. At lower biases (e.g., $V_G$ = 0.9 V, 1.2 V ), the model captures sharp, skewed distributions, while at higher voltages (e.g., $V_G$ = 1.6 V, 1.7 V), it accurately reproduces broader and even bimodal behavior. These results along with the high $R^2$ value of 0.92 confirm that the embedding-enhanced MDN not only learns the stochastic nature of FeFETs but also generalizes across operating conditions and device instances.

Together, these findings highlight the strength of the proposed approach in modeling real-world device variability without relying on the explicitly extracted physical parameters. Importantly, the model captures both cycle-to-cycle stochasticity and device-to-device variation, providing a comprehensive probabilistic prediction capability. Rather than relying on deterministic outputs or predefined corner models, the framework generates a distribution of possible outcomes rooted in data-driven uncertainty quantification. This enables designers to simulate a spectrum of behaviors that more accurately reflect real-world scenarios.

## 6. Conclusion

Building on a successfully demonstrated framework, our model incorporates key extensions that address prior limitations and broaden its applicability to FeFETs. In this work, we proposed a probabilistic modeling framework for FeFETs that captures both cycle-to-cycle and device-to-device variability using a device-specific embedding layer to learn latent physical characteristics. With an $R^2$ of 92% (root mean squared), our method has proven effective in capturing the inherent variations in FeFET behavior. Sampling from the learned embedding space enables the generation of synthetic device instances, providing a data-driven approach to variability-aware simulation. Furthermore, the use of C∞-continuous activation functions ensures compatibility with compact modeling requirements, allowing for smooth and differentiable outputs suitable for integration into circuit design tools. While not yet implemented as a full compact model, the architecture is well-suited for integration into compact modeling workflows and provides a strong foundation for future circuit-level simulation and design. Looking ahead, incorporating additional input features such as temperature, cycle count, or device structure may further enhance the accuracy and robustness of FeFET modeling. Tuning the number of mixture components could also improve the balance between model expressiveness and stability. Beyond compact modeling, this framework opens opportunities for broader FeFET-related applications, including reliability analysis, statistical variability studies, and digital twin development for predictive diagnostics. These directions will help extend the model's utility across diverse FeFET use cases and accelerate the design of next-generation non-volatile memory and computing systems.

# References

[1] G. Gildenblat, "Compact modeling," *Netherlands: Springer,* 2010.

[2] S. Dünkel *et al.*, "A FeFET based super-low-power ultra-fast embedded NVM technology for 22nm FDSOI and beyond," in *2017 IEEE International Electron Devices Meeting (IEDM)*, 2017: IEEE, pp. 19.7. 1-19.7. 4.

[3] A. Dasgupta *et al.*, "BSIM compact model of quantum confinement in advanced nanosheet FETs," *IEEE Transactions on Electron Devices,* vol. 67, no. 2, pp. 730-737, 2020.

[4] M. Reuter *et al.*, "Machine Learning Based Compact Model Design for Reconfigurable FETs," *IEEE Journal of the Electron Devices Society,* 2024.

[5] Y. Wang, Y. Zhang, E. Deng, J.-O. Klein, L. A. Naviner, and W. Zhao, "Compact model of magnetic tunnel junction with stochastic spin transfer torque switching for reliability analyses," *Microelectronics Reliability,* vol. 54, no. 9-10, pp. 1774-1778, 2014.

[6] E. Becle, P. Talatchian, G. Prenat, L. Anghel, and I.-L. Prejbeanu, "Fast behavioral VerilogA compact model for stochastic MTJ," in *ESSDERC 2021-IEEE 51st European Solid-State Device Research Conference (ESSDERC)*, 2021: IEEE, pp. 259-262.

[7] J. Wang, Y.-H. Kim, J. Ryu, C. Jeong, W. Choi, and D. Kim, "Artificial neural network-based compact modeling methodology for advanced transistors," *IEEE Transactions on Electron Devices,* vol. 68, no. 3, pp. 1318-1325, 2021.

[8] L. Zhang and M. Chan, "Artificial neural network design for compact modeling of generic transistors," *Journal of Computational Electronics,* vol. 16, pp. 825-832, 2017.

[9] J. Hutchins *et al.*, "A generalized workflow for creating machine learning-powered compact models for multi-state devices," *IEEE Access,* vol. 10, pp. 115513-115519, 2022.

[10] A. Zeumault, S. Alam, Z. Wood, R. J. Weiss, A. Aziz, and G. S. Rose, "TCAD modeling of resistive-switching of HfO2 memristors: Efficient device-circuit co-design for neuromorphic systems," *Frontiers in Nanotechnology,* vol. 3, p. 734121, 2021.

[11] S. George *et al.*, "Device circuit co design of FEFET based logic for low voltage processors," in *2016 IEEE Computer Society Annual Symposium on VLSI (ISVLSI)*, 2016: IEEE, pp. 649-654.

[12] M. Hoffmann *et al.*, "Antiferroelectric negative capacitance from a structural phase transition in zirconia," *Nature communications,* vol. 13, no. 1, p. 1228, 2022.

[13] N. Shukla *et al.*, "Ag/HfO 2 based threshold switch with extreme non-linearity for unipolar cross-point memory and steep-slope phase-FETs," in *2016 IEEE International Electron Devices Meeting (IEDM)*, 2016: IEEE, pp. 34.6. 1-34.6. 4.

[14] C. Roemer *et al.*, "Physics-based DC compact modeling of Schottky barrier and reconfigurable field-effect transistors," *IEEE Journal of the Electron Devices Society,* vol. 10, pp. 416-423, 2021.

[15] W. Ni, Z. Dong, B. Huang, Y. Zhang, and Z. Chen, "A physic-based explicit compact model for reconfigurable field-effect transistor," *IEEE Access,* vol. 9, pp. 46709-46716, 2021.

[16] C. Wang, P. Nulty, and D. Lillis, "A comparative study on word embeddings in deep learning for text classification," in *Proceedings of the 4th international conference on natural language processing and information retrieval*, 2020, pp. 37-46.

[17] H. Mulaosmanovic, E. T. Breyer, S. Dünkel, S. Beyer, T. Mikolajick, and S. Slesazeck, "Ferroelectric field-effect transistors based on HfO2: a review," *Nanotechnology,* vol. 32, no. 50, p. 502002, 2021.

[18] T. Hastie, R. Tibshirani, and J. Friedman, "The elements of statistical learning," ed: Springer series in statistics New-York, 2009.

[19] G. J. McLachlan and K. E. Basford, *Mixture models: Inference and applications to clustering.* M. Dekker New York, 1988.

[20] J. Hutchins, S. Alam, D. S. Rampini, B. G. Oripov, A. N. McCaughan, and A. Aziz, "Machine learning-powered compact modeling of stochastic electronic devices using mixture density networks," *Sci Rep,* vol. 14, no. 1, p. 6383, 2024.

[21] A. Radhakrishnan, D. Beaglehole, P. Pandit, and M. Belkin, "Mechanism for feature learning in neural networks and backpropagation-free machine learning models," *Science,* vol. 383, no. 6690, pp. 1461-1467, 2024.

[22] M. Abadi, "TensorFlow: learning functions at scale," in *Proceedings of the 21st ACM SIGPLAN international conference on functional programming*, 2016, pp. 1-1.

[23] F. Chollet, *Deep learning with Python*. simon and schuster, 2021.